\documentclass[lettersize, journal]{IEEEtran}
\IEEEoverridecommandlockouts                               

\usepackage{setspace}
\usepackage{subfiles}
\usepackage{subcaption}
\usepackage{algorithmic}
\usepackage{algorithm}
\usepackage{mathptmx}
\usepackage{graphicx} 
\usepackage{multirow}
\usepackage{soul}
\usepackage{diagbox}
\usepackage{array}
\usepackage{amsmath, amssymb, amsfonts}
\usepackage{todonotes}
\usepackage[colorlinks=true, allcolors=blue]{hyperref}
\usepackage{booktabs}

\usepackage{textcomp}
\usepackage{stfloats}
\usepackage{url}
\usepackage{verbatim}

\usepackage{xcolor}

\usepackage{cite}
\usepackage{subcaption}
\usepackage{comment}
\usepackage[justification=raggedright,singlelinecheck=false]{caption} %To left-align table captions in IEEEtran (default: center) 

\title{Multiclass Semantic Segmentation of Wildland Fire Images Using Context-Aware Centralized Copy-Paste Data Augmentation}

\author{Joon Tai Kim$^{1}$, Nishanth Kunchala$^{2}$, Vishv Patel$^{3}$, Tianle Chen$^{4}$, Ziyu Dong$^{5}$, Daniel Ospina Acero$^{6}$, Roger Williams$^{7}$, and Mrinal Kumar$^{8}$%
\thanks{* This work is supported by the National Science Foundation  under Grant No. 2132798 under the NRI 3.0: Innovations in Integration of Robotics Program.}%
\thanks{$^{1}$Joon Tai Kim is with the Mechanical and Aerospace Engineering Department, The Ohio State University, Columbus, OH 43210, USA 
        {\tt\small (email: kim.2209@osu.edu)}.}%
\thanks{$^{2}$Nishanth Kunchala is with the Mechanical and Aerospace Engineering Department, The Ohio State University, Columbus, OH 43210, USA 
        {\tt\small (email: kunchala.2@osu.edu)}.}%
\thanks{$^{3}$Vishv Patel is with the Mechanical and Aerospace Engineering Department, The Ohio State University, Columbus, OH 43210, USA 
        {\tt\small (email: patel.4644@osu.edu)}.}%
\thanks{$^{4}$Tianle Chen is with the Department of Computer Science, Boston University, Boston, MA 02215, USA
        {\tt\small (email: tianle@bu.edu)}. This work was done while at The Ohio State University (OSU).}%
\thanks{$^{5}$Ziyu Dong is with the Department of Wildland Resources, Utah State University, Logan, UT 84322, USA
     {\tt\small (email: ziyu.dong@usu.edu)}. This work was done while at The Ohio State University (OSU).}%
\thanks{$^{6}$Daniel Ospina Acero is with the Department of Electronic and Telecommunications Engineering, Universidad de Antioquia, Medell\'in, Colombia                  {\tt\small (email: daniel.ospina3@udea.edu.co)}.}
\thanks{$^{7}$Roger Williams is with the School of Environment and Natural Resources, The Ohio State University, Columbus, OH 43210, USA
        {\tt\small (email: williams.1577@osu.edu)}.}%
\thanks{$^{8}$Mrinal Kumar is with the Mechanical and Aerospace Engineering Department, The Ohio State University, Columbus, OH 43210, USA 
        {\tt\small (email: kumar.672@osu.edu)}.}%
}

\begin{document}

\maketitle

% Abstract
\begin{abstract}
Producing accurate annotations for deep learning based image segmentation is both costly and labor intensive. This challenge is especially evident in wildland fire applications, where accurately labeled datasets are scarce due to the difficulty of collecting and annotating dynamic fire scenes. To address this problem, our previous work introduced the Centralized Copy-Paste Data Augmentation (CCPDA) method for semantic segmentation of wildland fire imagery, which generates artificial training samples by randomly pasting fire clusters from source images onto target images. However, random placement can produce contextually unrealistic scenes, such as fire burning on asphalt. In this paper, we present a context-aware strategy designed specifically to improve data quality and realism in small multiclass wildland fire datasets, ensuring that augmented samples remain contextually meaningful. The proposed method restricts fire placement to semantically valid target regions and selects the location whose \textit{Ash–Vegetation} composition most closely matches the source context. This approach preserves existing fire regions in the target image, prevents unrealistic placements, and maintains contextual accuracy by generating images that resemble real wildland fire scenes. We evaluate the Context-Aware CCPDA strategy through numerical analysis and comparisons with other augmentation methods by a weighted sum-based multi-objective optimization (MOO) approach. The results confirm that the context-aware data augmentation strategy leads to improved segmentation performance and contextual realism, outperforming other augmentation procedures.
\end{abstract} 

% INTRODUCTION
\section{Introduction}

\subsection{Wildland Fire and the Role of Unmanned Aerial Systems} \label{Introduction: Wild_Fire}

Wildland fire is an ecological process that plays a crucial role in the global carbon cycle~\cite{bowman2009} and the preservation of fire-dependent ecosystems through regulation of vegetation structure and biodiversity~\cite{allen2002}. At the same time, wildfires along the wildland-urban interface (WUI) have grown in size and destructiveness over the past decade, with severe impacts on human life and property~\cite{radeloff2023rising, wang2021economic}. This trend is driven in part by changing climate patterns, other related extreme weather events, and human activity. In particular, fire exclusion policies implemented over the past century have led to an imbalance between live and dead vegetation, causing deterioration of forest health and creating dangerous levels of fuel buildup along the so-called wildland-urban interface~\cite{brown2023climate, ford2024wildland}. Frequent and large wildfires that threaten human communities have placed considerable stress on wildland fire management personnel and resources. Therefore, early detection of wildfires has become critical, especially in high-risk wildfire regions~\cite{harkat2023fire}. In this context, unmanned aerial systems (UAS) provide a means of real-time situational awareness, allowing early detection and continuous surveillance of active wildfires~\cite{duangsuwan2023accuracy, kim2015real, kim2026ccpda}.

The use of thermal sensors on UAS, including infrared (IR) systems, has proven effective for detecting fire fronts in complex wildfire environments with heavy smoke and dense vegetation. For instance, Kim et al. (2026) utilized a UAS equipped with an IR sensor to map active fire front boundaries and derive estimates of fire rate of spread~\cite{kim2026wildland}. Due to its accessibility, RGB imagery has also been widely used in wildfire research, where collected data are processed by vision-based algorithms, such as the multicolor thresholding method, to distinguish fire pixels from non-fire pixels~\cite{dang2019aerial}. 

Despite advances in aerial sensing, the effective use of UAS wildfire imagery for detailed analysis of wildfire behavior and reliable segmentation remains limited~\cite{kim2026ccpda}. In a related image-based fire detection study, Li and Zhao (2020) evaluated CNN-based object detectors, including Faster R-CNN, R-FCN, and YOLOv3, for fire and smoke detection. However, their study focused on bounding-box detection in mixed indoor and outdoor imagery rather than pixel-wise segmentation of realistic wildland fire scenes~\cite{li2020image}. Meanwhile, satellite-based methods developed for mapping large-scale burned areas do not translate well to high-resolution localized fire scenes~\cite{singh2025beyond}. Together, these limitations highlight a gap in reliable methods for pixel-wise segmentation of high-resolution wildland fire imagery. Consequently, there continues to be a need for robust segmentation algorithms specifically tailored to wildland fire conditions and suitable for practical deployment in operational firefighting contexts~\cite{kim2026ccpda}.  

\subsection{Deep Learning–Based Wildfire Semantic Segmentation} \label{Introduction: DL Based Seg}

In recent years, the deep-learning network U-Net~\cite{ronneberger2015u}, which was originally developed for biomedical image segmentation, has been adopted in other domains, including urban-scene understanding~\cite{solanki2023unet_road} and agricultural image analysis~\cite{sahin2023segmentation}. These studies demonstrate U-Net’s strong generalization ability for pixel-wise segmentation tasks, motivating its use in wildfire image segmentation. Rashkovetsky et al.~\cite{rashkovetsky2021wildfire} used a modified U-Net architecture to perform semantic segmentation of areas affected by wildfires using multisensor satellite imagery from Sentinel-1, Sentinel-2, Sentinel-3, Terra, and Aqua. By constructing a large and annotated dataset from the CAL FIRE perimeter database, the authors showed that combining information from multiple satellite sensors leads to more accurate wildfire segmentation, particularly under cloudy conditions. Other fully supervised neural network models, including FlameTransNet~\cite{chen2023FlameTransNet} and FBC-ANet~\cite{zhang2023FBC-ANet}, have also been proposed for UAS-based semantic segmentation of wildfire imagery. Chen et al. proposed FlameTransNet, which integrates Transformer blocks and a CBAM attention module, combining channel-wise and spatial attention, to improve binary flame segmentation of wildfire images. In contrast, Zhang et al. developed FBC-ANet, an Xception-based encoder-decoder model that improves boundary localization and employs a Contextual Information Awareness strategy to capture features at multiple spatial resolutions, allowing more reliable segmentation of UAV wildfire images.

A significant feature of the above-mentioned studies is that they only concern binary segmentation (\textit{Fire} vs. \textit{Non-fire}). While such an approach is useful for flame localization, it does not help resolve vegetation (\textit{fuel}, representing areas at risk for fire spread), ash (\textit{non-burnable}, representing already-burned regions), and other contextual elements that are crucial for decision-making in wildland fire management. Thus, multiclass labeled datasets such as BURN 1~\cite{BURN01} and BURN 2~\cite{BURN02}, which include the \textit{Vegetation}, \textit{Ash}, and \textit{Background} classes in addition to the \textit{Fire} class, provide a more detailed and operationally relevant representation of the wildfire environment. In particular, the BURN 2 dataset is constructed using the Context-Aware CCPDA strategy presented in this paper, which leverages the spatial statistics of neighboring pixels to generate more realistic synthetic samples.

\subsection{Limitations of Existing Wildfire Datasets} \label{Introduction: Data Limitations}

Although deep learning models such as the U-Net have provided reliable segmentation results in various applications, their success heavily depends on access to large, diverse, and well-annotated datasets~\cite{kim2026ccpda}. In wildland fire imagery, accurate labeling is challenging due to occlusion by smoke, the dynamic nature of flames, and the above-described need for multiclass representation of the wildfire environment. Most publicly available datasets, such as FLAME~\cite{shamsoshoara2020dataset} and FLAME 2~\cite{hopkins2022dataset}, are limited to binary segmentation (\textit{Fire} vs. \textit{Non-fire}) and therefore cannot distinguish other classes of interest (e.g., \textit{Ash} and \textit{Vegetation}) that are critical for wildland firefighting and resource management. For example, accurate segmentation of fuel (\textit{Vegetation}) is essential for estimating fire intensity and rate of spread. Similarly, identifying ash regions helps with the mapping of already-burned zones and the assessment of fire containment needs during an active burn. Our prior work~\cite{kim2026ccpda} proposed the CCPDA method to address this challenge by mitigating data scarcity and improving segmentation with limited labeled training data, using a strategy that copied only central fire regions onto new images while omitting error-prone boundary pixels; this concept is elaborated in Section~\ref{Introduction: Data Aug} next. 

\subsection{Data Augmentation} 
\label{Introduction: Data Aug}

Data augmentation is widely used to enhance model generalization when labeled data is scarce or imbalanced. Early work in this area relied on simple geometric transformations, such as random cropping and horizontal flipping, to generate additional training samples and improve model robustness~\cite{krizhevsky2012imagenet}. More recent methods, such as AutoAugment~\cite{cubuk2019autoaugment} and RandAugment~\cite{cubuk2020randaugment}, introduced automated policy search approaches that leverage reinforcement learning techniques and randomized exploration to identify effective augmentation strategies. For semantic segmentation tasks, non-context-aware copy-paste augmentation methods such as “Cut, Paste and Learn”~\cite{dwibedi2017cut} and “Simple Copy-Paste”~\cite{ghiasi2021simple} generate additional training samples by pasting labeled object masks into new images, thus improving the robustness of segmentation. In wildfire image segmentation, conventional augmentation methods, including rotations, flips, shifts, scaling, and chromatic distortion, have been used to expand training datasets~\cite{Khryashchev2020sat_unet}. Chen et al. introduced an adaptive copy-paste scheme in FlameTransNet that uses confidence-based sampling to paste flame pixels onto low-confidence training images, although the pasted flames are not guided by the semantic context of the target image~\cite{chen2023FlameTransNet}.

Limited availability of labeled wildfire data motivated our previous work~\cite{kim2026ccpda}, which introduced Centralized Copy-Paste Data Augmentation (CCPDA). This data augmentation approach improved the segmentation performance of wildfire imagery by placing only the core regions of fire clusters on new images, while excluding boundary pixels prone to labeling errors. Although the CCPDA method performed better than other augmentation approaches (e.g., the 10\% erosion approach achieved the lowest \textit{Fire} false-negative rate of 5.14\%), it raised a key concern regarding whether training on contextually inaccurate representations of fire is appropriate, e.g., fire pasted onto roads. Here, erosion refers to the removal of boundary pixels from fire clusters to preserve only their central high-confidence regions. The original CCPDA method, hereafter referred to as the Non-Context-Aware CCPDA method, could randomly paste fire clusters onto existing fire regions or non-burnable areas (e.g., on people, water, or parking lots), reducing the number of trainable \textit{Fire} pixels and producing unrealistic training samples. This limitation naturally prompted the question of whether fire classification can be improved if the augmentation of wildland fire data is carried out in a contextually correct manner. By ``contextually correct'', we refer to placing fire clusters in semantically valid target regions whose local \textit{Ash–Vegetation} composition aligns with that of the source cluster’s neighboring pixels. This concept is further detailed in Sections~\ref{Introduction: Context-Aware} and~\ref{Introduction: Objective}.

\subsection{Context-Aware Data Augmentation} 
\label{Introduction: Context-Aware}

Non-contextual copy-paste data augmentation methods, which paste objects randomly without considering spatial or physical context, have been widely adopted and shown to improve the robustness of image segmentation models. However, to improve the realism of augmented images, more context-aware alternatives have also been explored. For example, InstaBoost~\cite{fang2019instaboost} uses a location probability map to identify feasible paste locations based on local appearance similarity, rather than relying on purely random placement. More recently, Guo et al. (2025) showed that random placements, such as those used in “Simple Copy-Paste''~\cite{ghiasi2021simple}, can still produce unrealistic and semantically incorrect images, and therefore proposed the Context-Aware Copy-Paste (CACP) method to address this limitation~\cite{guo2025cacp}. Instead of placing objects randomly, CACP employs BERT-based semantic matching to align pasted objects with contextually compatible scenes, thereby improving the realism of the augmented images. For example, the non-contextual method may paste an ostrich onto a soccer field, creating an implausible training sample, whereas the context-aware method restricts placement to scenes that better reflect the object’s natural environment. Context-aware augmentation is especially important in wildfire applications, where unrealistic fire placement can reduce the validity of training data. Accordingly, we propose the Context-Aware CCPDA method, which directs the placement of fire clusters using semantic placement constraints and the \textit{Ash–Vegetation} composition of neighboring pixels.
 
\subsection{Objectives and Contributions} 
\label{Introduction: Objective}

This work builds on prior research on the Non-Context-Aware CCPDA method~\cite{kim2026ccpda} for wildland fire image segmentation with limited labeled data. Two central questions are posed: (1) How to mathematically frame the notion of \textit{contextually correct} augmentation of fire samples? (2) To what extent does the Context-Aware CCPDA method improve wildland fire segmentation compared to the Non-Context-Aware CCPDA approach? The ultimate objective of this study is to improve the semantic segmentation of wildland fire imagery under limited labeled data conditions, with particular emphasis on reducing missed detections of the \textit{Fire} class in the segmentation output.

\begin{figure*}[!ht]
    \centering
    \captionsetup{justification=centering}
    \includegraphics[width=0.85\textwidth]{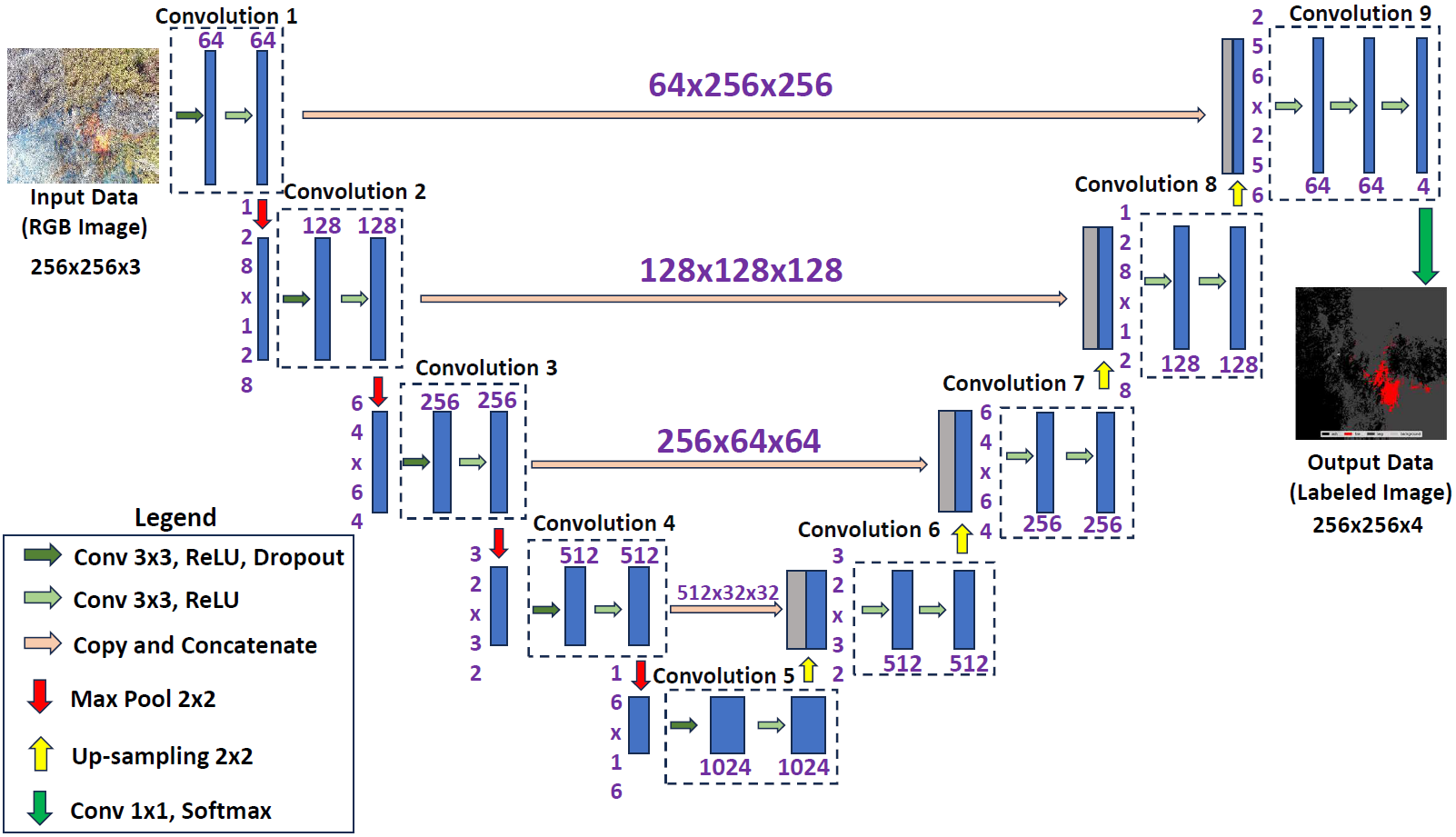}
    \caption{The U-Net architecture used as the baseline model for semantic segmentation of wildland fire images. This figure is reused from~\cite{kim2026ccpda}.}
    \label{fig:unet_structure}
\end{figure*}

For each fire cluster extracted with the Context-Aware CCPDA method, we compute the \textit{Ash–Vegetation} composition of the surrounding pixels within its bounding box. To determine a contextually compatible placement on the target image, we first identify valid candidate locations that satisfy semantic placement constraints, ensuring that pasted fire clusters do not overlap existing \textit{Fire} pixels or non-burnable \textit{Background} regions. Among these valid candidate locations, the optimal placement is selected by minimizing a cost function that quantifies the contextual difference as the Euclidean distance between the \textit{Ash–Vegetation} compositions of the source and target regions. We construct the BURN 2~\cite{BURN02} dataset from a set of 26 (\textit{non-augmented}) RGB images and their corresponding four-class ground-truth labels for \textit{Ash}, \textit{Fire}, \textit{Vegetation}, and \textit{Background}. Augmented samples are then generated from this image set using the proposed Context-Aware CCPDA method. The resulting dataset is used within a U-Net based semantic segmentation framework (see Figure~\ref{fig:unet_structure}). Finally, we adopt a weighted sum-based multi-objective optimization (MOO) strategy designed to reduce false negatives in the \textit{Fire} class, ensure accurate \textit{Vegetation} segmentation, and maintain high overall performance.

\noindent The contributions of this work are enumerated below:
\begin{enumerate}
\item We introduce the Context-Aware CCPDA method, a context-aware extension of the original Non-Context-Aware CCPDA method that improves semantic segmentation performance by guiding the placement of fire clusters from source images onto contextually compatible regions of target images according to local \textit{Ash-Vegetation} composition. This approach preserves contextual similarity between source and target regions, thereby producing more realistic augmented training samples. As detailed in Section~\ref{sec:Study01}, the Context-Aware CCPDA method achieves the lowest \textit{Fire} false-negative rate (\textit{Fire}-FNR) of 6.75\% and the highest $F(x)$ score of 0.8458 among the tested augmentation methods.

\item We present an expanded wildland fire image dataset by expanding the image set used to construct BURN 1~\cite{BURN01}, which was generated from 20 (\textit{non-augmented}) RGB images and their corresponding ground-truth labels using the Non-Context-Aware CCPDA method. In the present work, we add six RGB images and their corresponding ground-truth labels, collected during the same prescribed burn in central Ohio, expanding the image set from 20 to 26 (\textit{non-augmented}) image-label pairs. This expanded set is used to create the BURN 2 dataset with the proposed Context-Aware CCPDA framework. In particular, the six additional images include more complex background elements (e.g., people, vehicles, and parking areas) than the original 20 images, thereby increasing object diversity and establishing a more challenging benchmark for evaluating the Context-Aware CCPDA method against other augmentation strategies.

\item To determine whether the same erosion level provides the best performance for the Context-Aware CCPDA method before and after expansion of the original image set, we apply the method at erosion levels of 0\%, 10\%, 20\%, and 30\% to two image sets: the original 20-image set used to construct BURN 1 and the expanded 26-image set used to construct BURN 2. As detailed in Section~\ref{sec:Study02}, the 10\% erosion level achieves the lowest \textit{Fire}-FNR and highest $F(x)$ score for both image sets, with \textit{Fire}-FNR values of 4.00\% and 4.67\% and $F(x)$ scores of 0.8365 and 0.8460, respectively. These results demonstrate that the best-performing erosion level remains consistent even after expansion of the original annotated image set.

\item Finally, we demonstrate that the performance gains achieved by the Context-Aware CCPDA method are not limited to the U-Net architecture but extend across multiple deep-learning-based semantic segmentation architectures, as shown in Section~\ref{sec:Study03}. The Context-Aware CCPDA method achieves the lowest \textit{Fire}-FNR and highest $F(x)$ score among the tested augmentation methods for FCN~\cite{long2015fcn}, SegNet~\cite{badrinarayanan2017segnet}, and DeepLabV3+~\cite{chen2018deeplabv3+}, with \textit{Fire}-FNR values of 21.72\%, 7.78\%, and 9.56\% and $F(x)$ scores of 0.7211, 0.8311, and 0.8146, respectively.
\end{enumerate}

% U-Net Figure

% METHODOLOGY
\section{Methodology}
% Intro Paragraph 
In this section, we present the Context-Aware CCPDA augmentation method and introduce the framework for comparative analysis with other augmentation strategies. 

\subsection{Data Collection and Preprocessing} 
\label{Method: Data Collection and Preprocessing}

The dataset used in this work was collected during a prescribed burn at Larry R. Yoder Prairie on the Marion campus of the Ohio State University (OSU) in Central Ohio. RGB images were acquired using a DJI Mavic 2 drone operated at altitudes ranging from 4 to 128 meters, with the camera oriented vertically downward to capture the bird’s eye view. Data collection occurred under clear lighting conditions on a sunny autumn afternoon between 1:00 and 2:30 PM. From these flights, 26 images exhibiting varying levels of smoke were selected for our analysis.

\begin{figure}[!h]
    \centering
    \captionsetup{justification=centering}
    \includegraphics[width=0.47\textwidth]{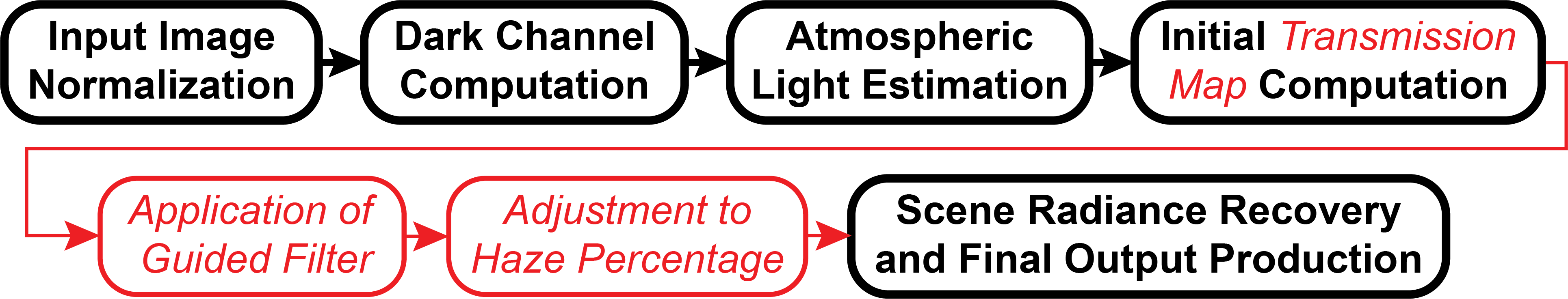}
    \caption{Flowchart for smoke dehazing method. This figure is reused from~\cite{kim2026ccpda}.}
    \label{fig:dehaze_flowchart}
\end{figure}

\begin{figure}[!h]
    \centering
    \captionsetup{justification=centering}
    \begin{subfigure}[b]{0.235\textwidth}
        \centering
        \includegraphics[width=\textwidth]{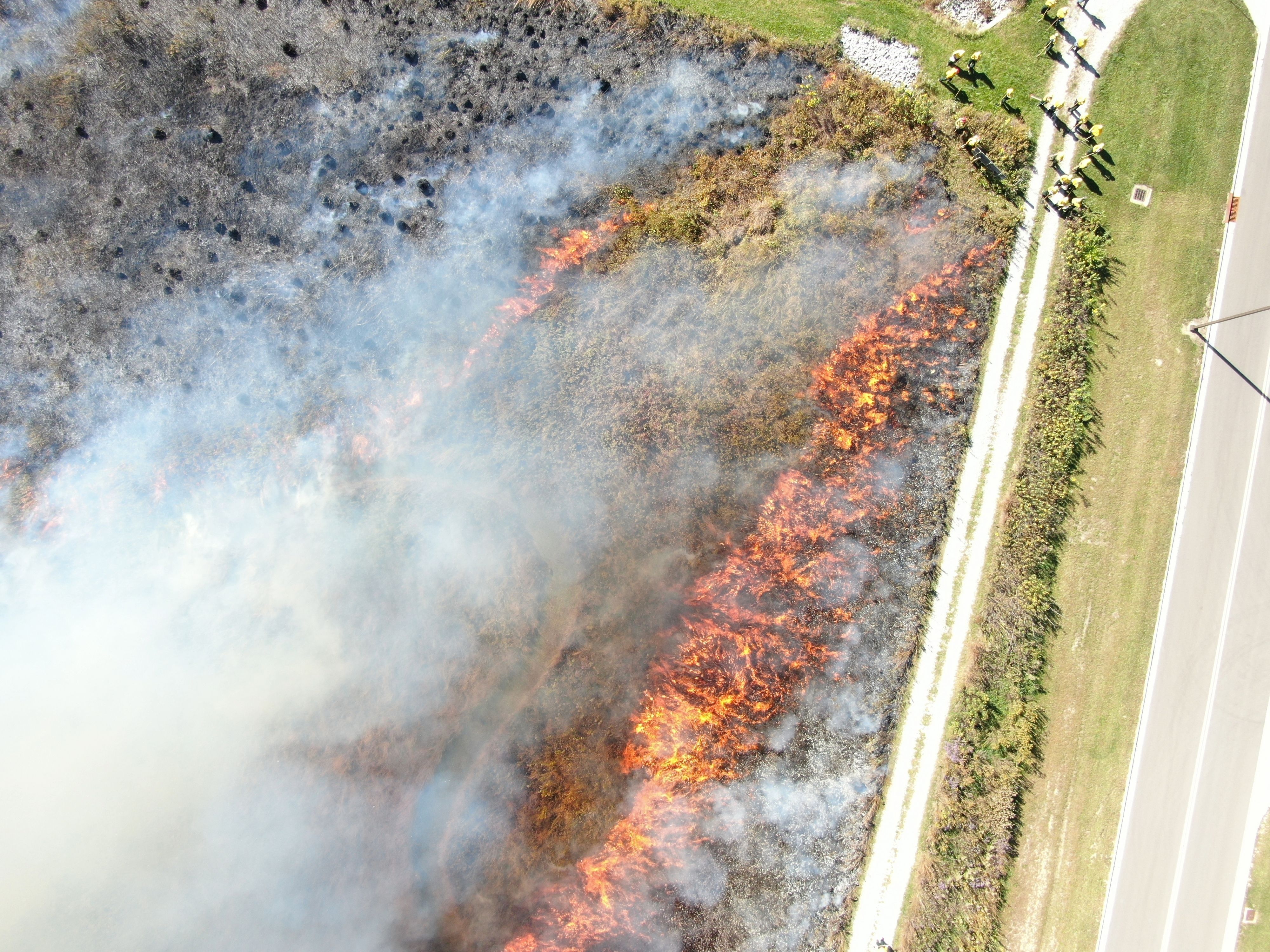}
        \captionsetup{justification=centering}
        \caption{}
        \label{fig:dehazing_a}
    \end{subfigure}
    \hfill
    \begin{subfigure}[b]{0.235\textwidth}
        \centering
        \includegraphics[width=\textwidth]{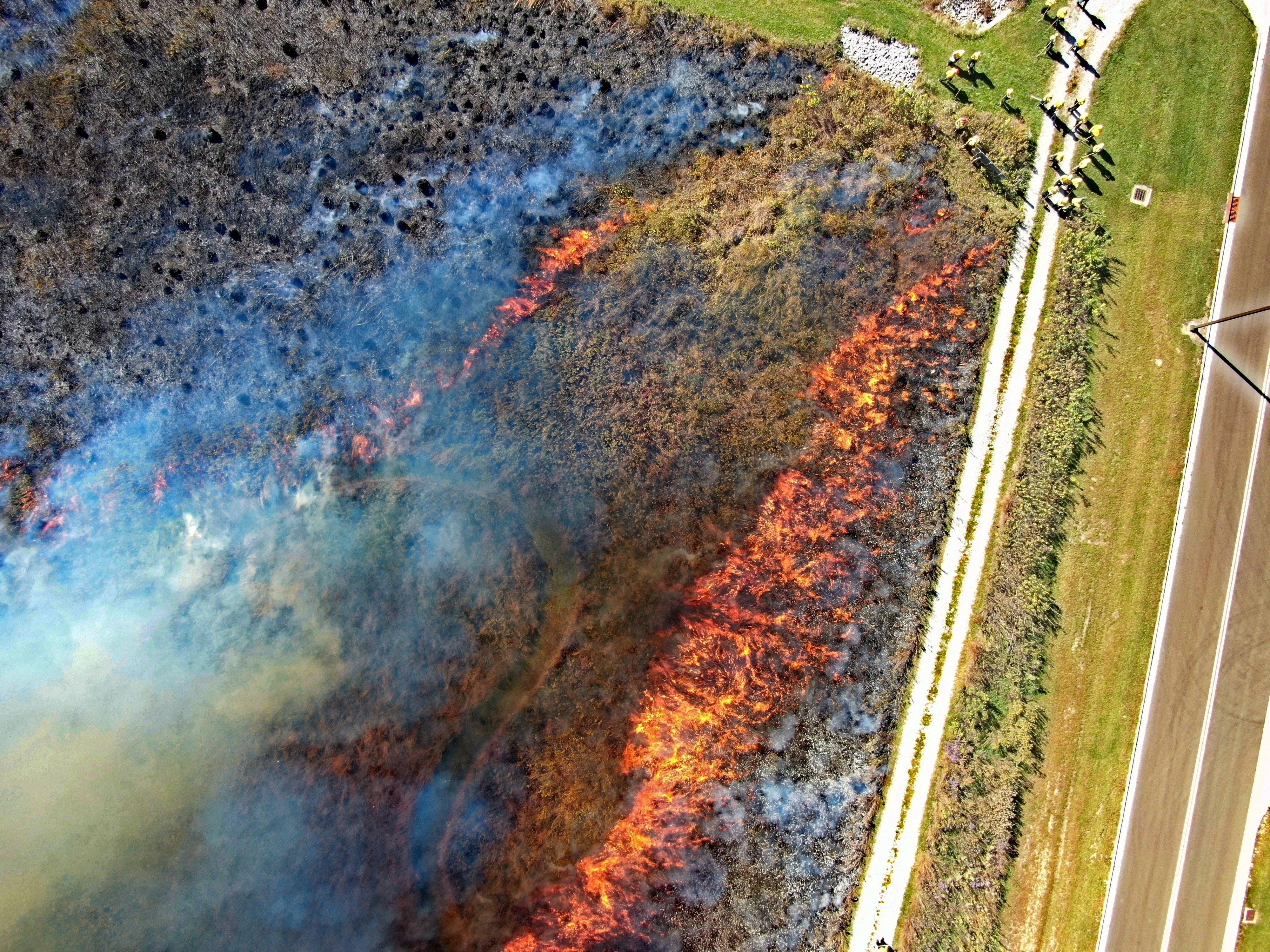}
        \captionsetup{justification=centering}
        \caption{}
        \label{fig:dehazing_b}
    \end{subfigure}
    \caption{A comparison of the original hazy RGB image (a) and the dehazed RGB image (b) achieved through a 95\% haze-reduction procedure.}
    \label{fig:dehaze_compare}
\end{figure}

To prepare images for use as a training dataset, we applied the dehazing algorithm proposed by He et al.~\cite{he2011dehaze, he2013guided}. This step is necessary because the smoke in RGB wildfire imagery can cause occlusion, making it difficult for the human eye to distinguish and accurately label the classes in smoke-covered regions. The smoke dehazing scheme utilized in this work is briefly described in Figure~\ref{fig:dehaze_flowchart}. A visual comparison between the original hazy image and the result of the dehazing process is shown in Figure~\ref{fig:dehazing_a} and Figure~\ref{fig:dehazing_b}, respectively, demonstrating the effectiveness of the 95\% haze-reduction setting. This setting corresponds to the parameter value used in He’s method~\cite{he2011dehaze}, where the tunable factor controls the degree of haze removal during transmission estimation.

As the final preprocessing step, we used ``ImageJ'' software with the Labkit plug-in to manually annotate the 26 RGB images and generate ground truth labels for four classes:  \textit{Ash}, \textit{Fire}, \textit{Vegetation}, and \textit{Background}. These annotations provide the basis for training deep learning segmentation models. It should be noted that in this study, smoke was classified as \textit{Background}. Although smoke provides information about fire presence and wind conditions, it was not treated as a separate semantic class in this study because the present framework focuses on surface-level \textit{Ash}, \textit{Fire}, \textit{Vegetation}, and \textit{Background} classes. For use in each augmentation process described below in Section~\ref{Method: Data Augmentation}, both the images and their corresponding labels are downscaled from $4000\times3000$ to $256\times256$ to satisfy the input requirements and computational constraints of the U-Net model.

% Dehaze Flowchart Figure 

% Dehaze Output Images 

\subsection{Data Augmentation} 
\label{Method: Data Augmentation}

\subsubsection{Non-Augmented Method} 
\label{Data Augmentation: Non-Augmented}

The non-augmented training dataset was used as a baseline to evaluate the benefit of data augmentation techniques. To ensure a balanced representation, the six newly annotated images were distributed equally across the original training, validation, and test sets, with two images added to each set. As a result, the final dataset split is \((8{+}\textbf{2}) : (2{+}\textbf{2}) : (10{+}\textbf{2})\) for training, validation, and test sets, respectively, where the first term in each pair indicates the original data and the second term (in bold) represents the newly added data. The test dataset remained non-augmented to serve as a control set to evaluate the accuracy of the model on untouched data. 

\subsubsection{Rotation-Based Method} 
\label{Data Augmentation: Rotation}

In this study, each image was rotated in 18-degree increments as a data augmentation strategy. This step size was chosen because rotating an image in 18-degree increments up to 360 degrees yields 20 augmented versions per image. For example, starting with 10 original non-augmented images in the training dataset, this process produced a total of 200 rotated samples. The final size of the dataset therefore matches that generated by both copy-paste approaches (described in Sections~\ref{Data Augmentation: Non-Context-Aware} and~\ref{Data Augmentation: Context-Aware}), ensuring consistency across augmentation methods. Each image was rotated starting at 5 degrees and incremented by 18 degrees to 360 degrees. To account for potential out-of-frame regions near the image boundaries, the rotated images were scaled by a factor of $1.66$ and cropped back to their original dimensions. 

\subsubsection{Non-Context-Aware Copy-Paste Method} 
\label{Data Augmentation: Non-Context-Aware}

The Non-Context-Aware CCPDA method of Ref.~\cite{kim2026ccpda} is used as the non-contextual data augmentation baseline for wildland fire image segmentation. In this approach, fire clusters from a source image are pasted onto a target image at random locations without considering the surrounding pixel composition. The dataset to be augmented is defined as
\begin{equation}
    \mathcal{D} = \{(I_i, M_i)\}_{i=1}^{n},
    \label{eq:dataset}
\end{equation}
where $I_i \in \mathbb{R}^{H \times W \times 3}$ represents the $i$-th RGB image in the dataset and $M_i \in \{\mathit{Ash},\ \mathit{Fire},\ \mathit{Vegetation},\ \mathit{Background}\}^{H \times W}$ denotes its corresponding multiclass segmentation mask. Next, to isolate fire regions for cluster extraction and subsequent morphological erosion, the multiclass mask $M_i$ is converted into a binary mask (\textit{Fire} vs.\ \textit{Non-fire}) $M_i^{\mathrm{bin}}$ as
\begin{equation} 
    M_i^{\mathrm{bin}}(p) =
    \begin{cases}
    1, & \text{if } M_i(p) = \mathit{Fire}\\
    0, & \text{otherwise}
    \end{cases},
    \label{eq:binary_fire_mask}
\end{equation}
where $p \in \{1,\ldots,H\}\times\{1,\ldots,W\}$ denotes a pixel location in the segmentation mask.

For each image $I_i$ and its corresponding binary mask $M_i^{\mathrm{bin}}$, the set of fire clusters is defined as
\begin{equation}
    S_i = \{s_{i,j}\}_{j=1}^{N_i},
    \label{eq:fire_clusters}
\end{equation}
where $s_{i,j}$ denotes the $j$-th fire cluster identified in $M_i^{\mathrm{bin}}$, and $N_i$ denotes the total number of fire clusters in $M_i^{\mathrm{bin}}$. To remove small noisy fragments, any fire cluster with $\lvert s_{i,j} \rvert < 100$ pixels is discarded. Although the non-context-aware copy-paste augmentation enhances dataset diversity and supports model generalization, labeling errors can occur along the boundaries of fire clusters. These boundary pixels are particularly prone to ambiguity because they may be partially obscured by smoke or visually similar to surrounding vegetation, which can introduce label noise during model training.

To mitigate this issue, the Non-Context-Aware CCPDA method applies morphological erosion to retain only the core regions of the fire clusters. Specifically, each fire cluster is eroded as
\begin{equation}
    \hat{s}_{i,j} = s_{i,j} \ominus K_x,
    \label{eq:fire_cluster_erosion}
\end{equation}
where $\ominus$ denotes morphological erosion and $K_x$ is an $x \times x$ square structuring element (a matrix of ones). Under this operation, a \textit{Fire} pixel is retained only if all pixels within the neighborhood defined by $K_x$ belong to the \textit{Fire} class. Otherwise, the pixel is removed. This operation removes boundary pixels from each fire cluster, retaining only the more reliable inner regions that are less prone to labeling ambiguity.

Next, to smooth the boundaries of the eroded fire clusters and reduce jagged edges introduced during erosion, we apply morphological dilation using a $5 \times 5$ square structuring element $K_5$:
\begin{equation}
    s'_{i,j} = \hat{s}_{i,j} \oplus K_5,
    \label{eq:fire_cluster_dilation}
\end{equation}
where $\oplus$ denotes the morphological dilation operator. Specifically, a pixel is assigned to the fire cluster if at least one pixel within the neighborhood defined by $K_5$ belongs to the \textit{Fire} class. Otherwise, the pixel remains \textit{Non-fire}. Although dilation expands the eroded fire clusters, it operates only on the pixels retained after erosion and therefore cannot necessarily reconstruct the original cluster boundaries.

Following the erosion and dilation steps, each refined fire cluster $s'_{i,j}$ is randomly rotated by an angle $\theta$ drawn from a uniform distribution, $\theta \sim \mathcal{U}(0,360^\circ)$. The rotated fire cluster is then pasted onto the target RGB image $I_k$ at a random location, where $k \in \{1,\ldots,n\}$ denotes the target-image index. The corresponding pixels in the multiclass target mask $M_k$ are updated at the same location. This produces the updated target image $I'_k$ and its corresponding multiclass mask $M'_k$.

\begin{figure}[t]
    \centering
    \captionsetup{justification=centering}
    \includegraphics[width=0.5\textwidth]{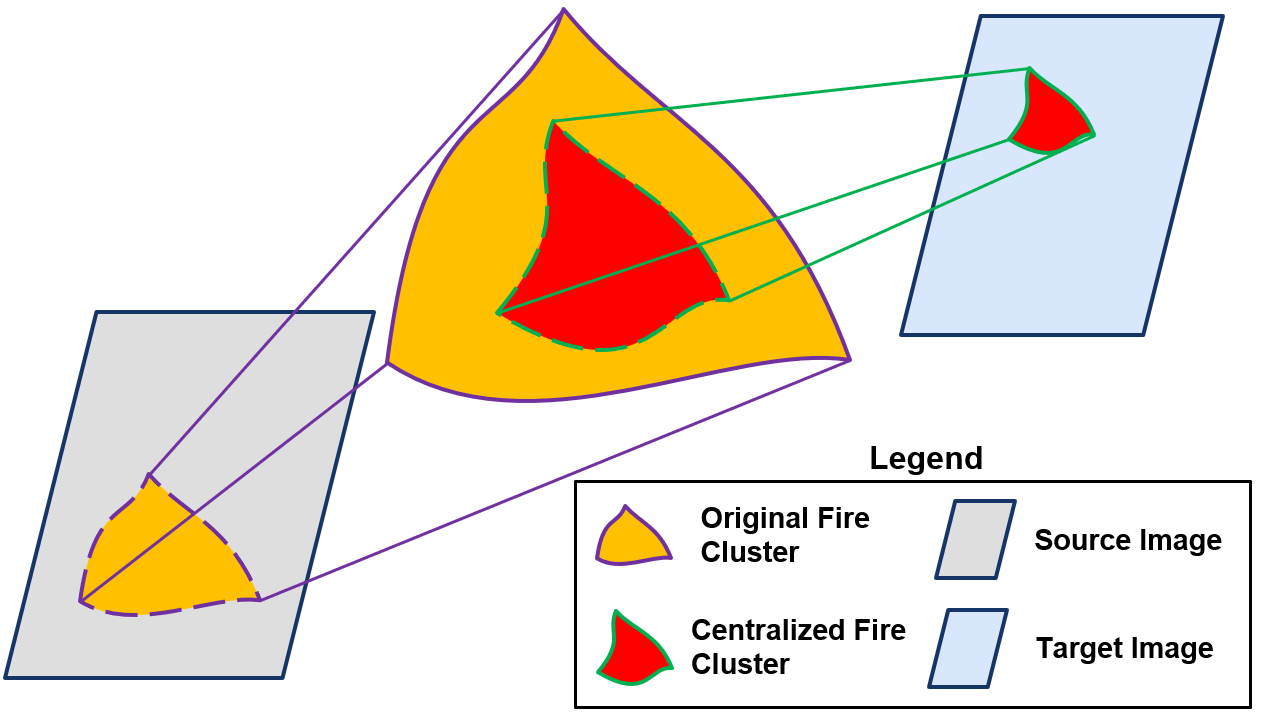}
    \caption{An illustration of the standard centralized copy-paste method for fire segmentation. The process involves: (1) identifying an original fire cluster (yellow) in the source image, (2) applying the centralized method to focus on the core fire area (red), and (3) pasting the centralized fire cluster onto a target image. This figure is reused from~\cite{kim2026ccpda}.}
    \label{fig:Copy-Paste-pipeline}
\end{figure}

Finally, the augmentation process is repeated for every source-target image pair, including self-pairings in which the source and target are the same image. Repeating this procedure for $r$ random seeds produces the augmented dataset,
\begin{equation}
    \mathcal{D}' = \{(I'_l, M'_l)\}_{l=1}^{n^2 r}.
    \label{eq:augmented_dataset}
\end{equation}
Here, $I'_l$ denotes the $l$-th resulting augmented image and $M'_l$ its corresponding segmentation mask, where $l \in \{1,\ldots,n^2r\}$ indexes the resulting augmented image-mask pairs. Although random placement increases training diversity, fire clusters can be pasted onto contextually implausible regions or overlap existing fire regions in the target image. Figure~\ref{fig:Copy-Paste-pipeline} illustrates the overall procedure.

% CCPDA Method Figure 

\subsubsection{Context-Aware Copy-Paste Method} 
\label{Data Augmentation: Context-Aware}

To generate semantically meaningful data augmentation, this paper introduces a context-aware copy-paste method that pastes fire clusters from source images onto realistic locations in target images. Named Context-Aware CCPDA, the method proceeds as follows. First, for each image-mask pair $(I_i, M_i)$, the corresponding binary mask (\textit{Fire} vs. \textit{Non-fire}) $M_i^{\mathrm{bin}}$ and the set of fire clusters $S_i$ are obtained as defined in Eqs.~\eqref{eq:binary_fire_mask} and \eqref{eq:fire_clusters}, respectively. For each fire cluster $s_{i,j}$, a bounding box $\mathcal{B}_{i,j}$ is generated around the cluster. Then, for each target image $I_k$ and its corresponding mask $M_k$, these bounding boxes are swept across $M_k$ in a sliding-window manner. 

For each source fire cluster, the following sequential evaluation is performed:
\begin{enumerate}
    \item \textbf{Semantic class composition} — for each fire cluster, quantify the proportions of \textit{Vegetation}, \textit{Ash}, \textit{Fire}, and \textit{Background} pixels within its bounding box $\mathcal{B}_{i,j}$;
    \item \textbf{Semantic placement constraint} — slide the source fire cluster bounding box $\mathcal{B}_{i,j}$ across the target mask $M_k$, evaluate each candidate location, and reject any location where the fire cluster mask overlaps existing \textit{Fire} pixels or non-burnable \textit{Background} pixels (e.g., roads, people);
    \item \textbf{Contextual similarity} — select the optimal candidate location according to Eq.~\eqref{eq:optimal-paste-loc} by minimizing the cost function in Eq.~\eqref{eq:ash-veg-cost}, defined as the Euclidean distance between the \textit{Ash-Vegetation} compositions of the source fire cluster bounding box $\mathcal{B}_{i,j}$ and each valid candidate region in the target mask $M_k$.
\end{enumerate}

The cost function used to evaluate contextual similarity is defined as
\begin{equation}
    J(x,y) =
    \sqrt{
    \left(a_{\mathrm{src}} - a_{\mathrm{trgt}}(x,y)\right)^2 +
    \left(v_{\mathrm{src}} - v_{\mathrm{trgt}}(x,y)\right)^2
    }.
    \label{eq:ash-veg-cost}
\end{equation}
In Eq.~\eqref{eq:ash-veg-cost}, $a_{\mathrm{src}}$ and $v_{\mathrm{src}}$ are the \textit{Ash} and \textit{Vegetation} percentages within the source fire cluster bounding box $\mathcal{B}_{i,j}$, while $a_{\mathrm{trgt}}(x,y)$ and $v_{\mathrm{trgt}}(x,y)$ represent the corresponding percentages within the valid candidate region at location $(x,y)$ in the target mask after the source fire cluster is temporarily placed at that location. Then, the optimal paste location is obtained by minimizing the cost function over the set of valid candidate locations:
\begin{equation}
    (x^*,y^*) =
    \underset{(x,y)\in\mathcal{C}_{i,j,k}}{\arg\min}\; J(x,y),
    \label{eq:optimal-paste-loc}
\end{equation}
where $\mathcal{C}_{i,j,k}$ is the set of valid candidate locations for placing the source fire cluster $s_{i,j}$, which is enclosed by bounding box $\mathcal{B}_{i,j}$, on the target mask $M_k$. The resulting coordinates $(x^*,y^*)$ define the optimal paste location.

\begin{figure*}[t]
    \centering
    \captionsetup{justification=centering}
    \includegraphics[width=\textwidth]{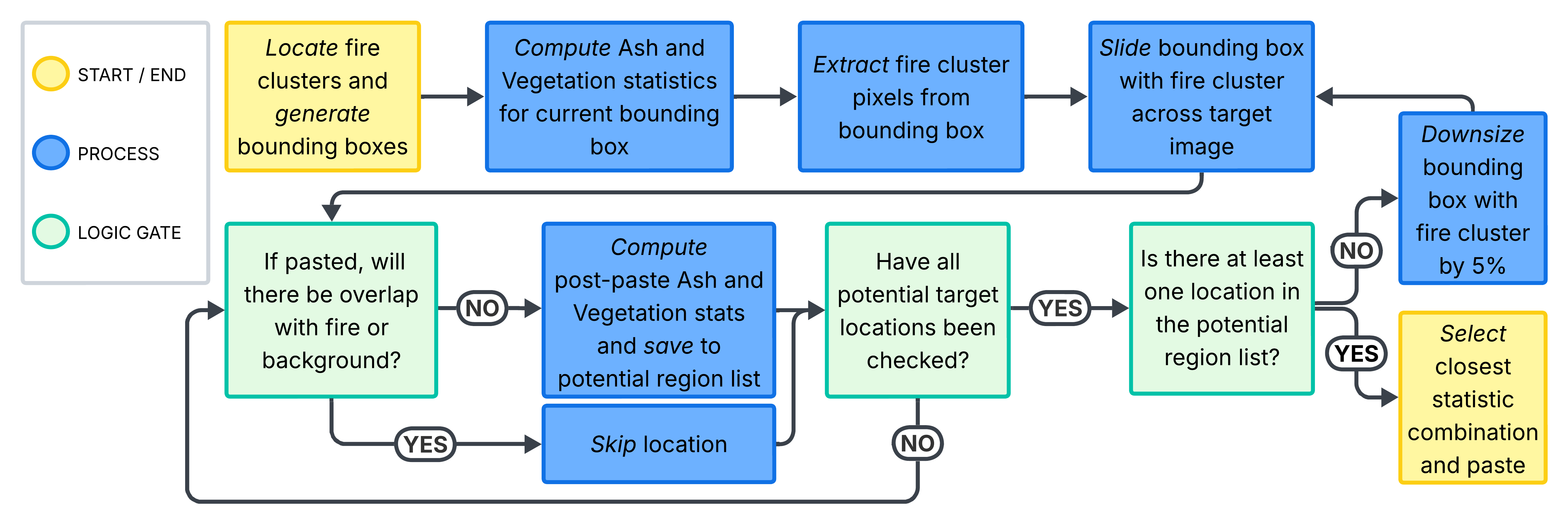}
    \caption{Flowchart illustrating the proposed Context-Aware CCPDA method. The process identifies fire clusters, computes \textit{Ash} and \textit{Vegetation} statistics, and evaluates candidate paste locations in the target image based on compositional similarity and semantic constraints. Finally, the optimal location is selected; if none is identified, the bounding box is adaptively downsized and the process repeated.}
    \label{fig:CACCPDA Flowchart}
\end{figure*}

If no paste location satisfies the semantic placement constraint, the source fire cluster and all non-fire pixels within its bounding box are uniformly scaled down by 5\%, and the search is repeated using the resized bounding box until at least one valid paste location is identified. If multiple optimal locations achieve the same minimum distance, one is randomly selected for pasting. It is important to note that the paste-location optimization is performed sequentially. Each selected placement updates the target image-mask pair, so the search at iteration $q+1$ is performed on the target image-mask pair produced at iteration $q$, where $q$ denotes the fire cluster placement iteration. A summary of the overall procedure is provided in Figure~\ref{fig:CACCPDA Flowchart}.

In contrast to the Non-Context-Aware CCPDA method (Section~\ref{Data Augmentation: Non-Context-Aware}), the Context-Aware CCPDA method prevents unrealistic fire placement onto the target image. By incorporating the surrounding environmental context into the augmentation process, this method produces contextually realistic fire placements, resulting in meaningful training samples. This context-aware augmentation strategy not only increases dataset diversity but also improves contextual accuracy, thereby strengthening the model’s ability to generalize to complex wildland fire scenes.

\begin{figure*}[!ht]
    \centering
    \captionsetup{justification=centering}
    \includegraphics[width=0.95\textwidth]{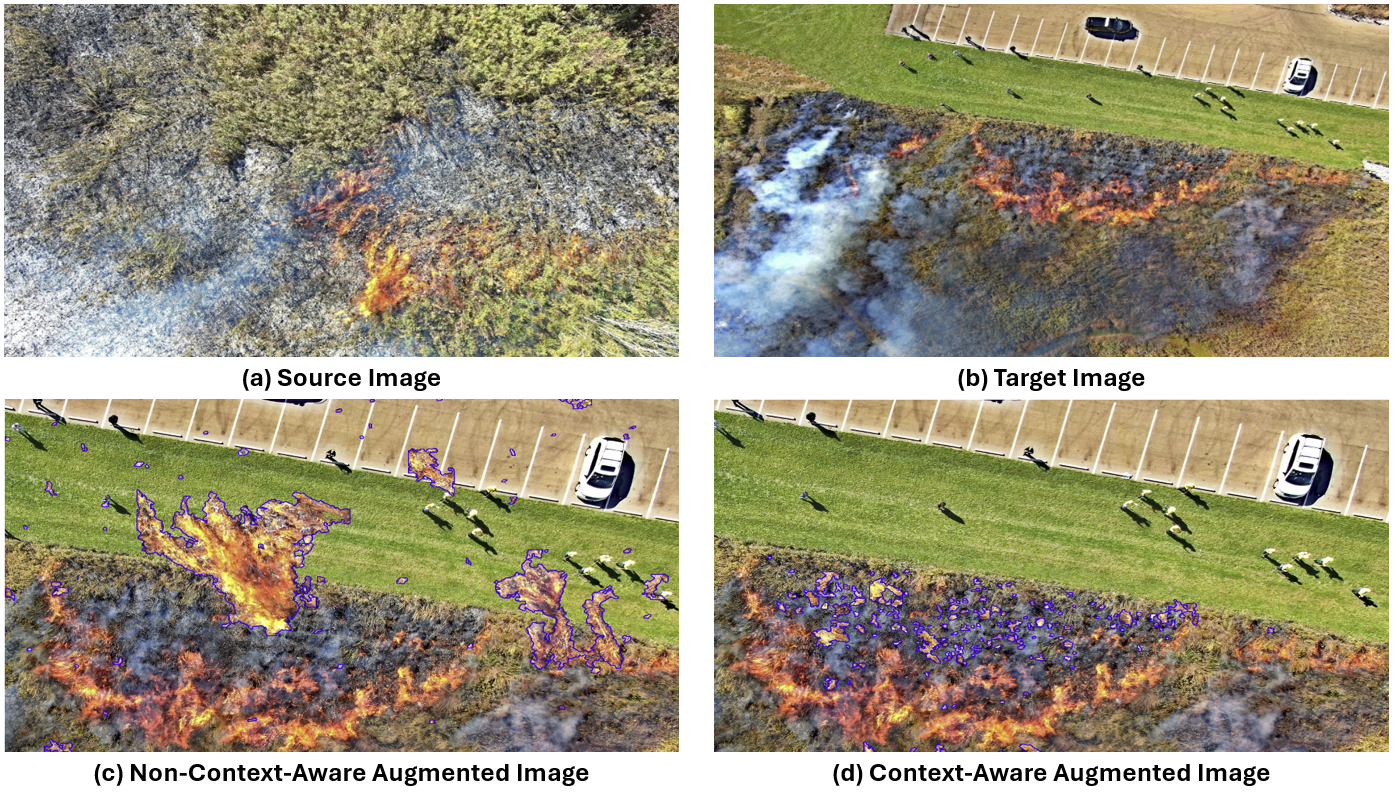}
    \caption{A visual comparison of Non-Context-Aware and Context-Aware CCPDA results for a sample source and target image pair. Sub-figures (a) and (b) show the original source and target images, respectively, while (c) and (d) display zoomed-in augmented images from the upper-center region of (b), generated using the Non-Context-Aware and Context-Aware CCPDA methods, respectively.}
    \label{fig:CCPDA Results Comparison}
\end{figure*}

As a demonstration of this method, the source and target images are shown in  Figures~\ref{fig:CCPDA Results Comparison}(a) and (b). The result of the non-context-aware method on this image pair is shown in Figure~\ref{fig:CCPDA Results Comparison}(c), illustrating that fire clusters are randomly pasted onto locations where fire cannot realistically occur, such as a parking lot or outside the prescribed burn area. In addition, the pasted fire clusters may overlap existing fire clusters in the target image, potentially overwriting high-quality ground-truth labels. In contrast, Figure~\ref{fig:CCPDA Results Comparison}(d) shows that the Context-Aware CCPDA method places fire clusters in locations where fire is expected to occur, such as vegetated regions within the prescribed burn area. This result further shows that the Context-Aware CCPDA method prevents overlap between pasted fire clusters and existing \textit{Fire} pixels in the target image.

\begin{figure}[t]
    \centering
    \captionsetup{justification=centering}
    \includegraphics[width=0.5\textwidth]{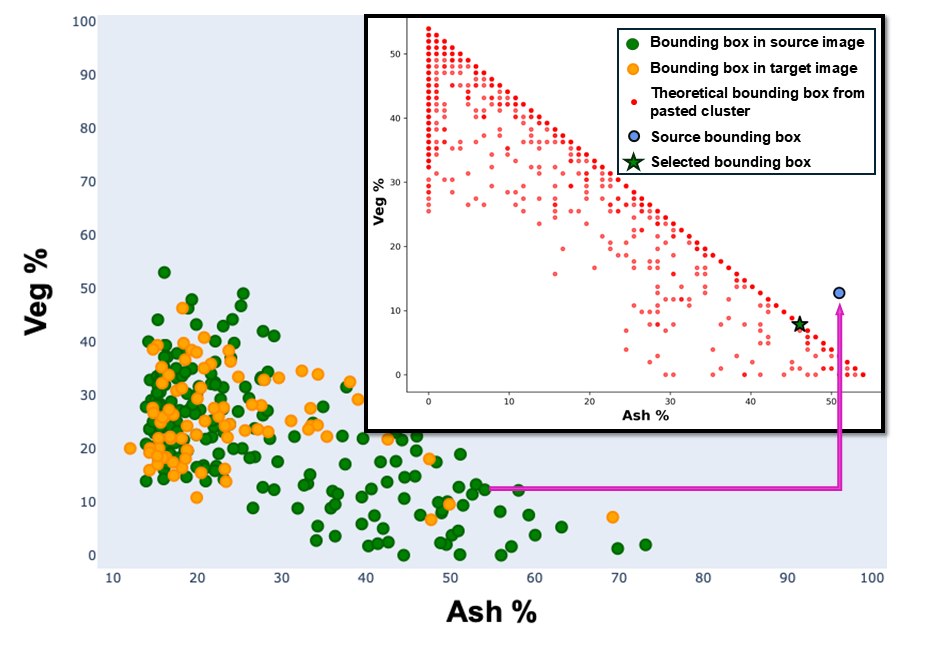}
    \caption{A visualization of compositional matching for wildland fire image data augmentation. Green and yellow data points represent \textit{Ash}–to-\textit{Vegetation} percentage distributions from the source and target images, respectively. The inset illustrates the context-aware paste-location selection process for a single source fire cluster, where candidate target bounding boxes (red points) are compared with the source bounding box (single blue point), and the target bounding box with the closest compositional match is selected (single green star).}
    \label{fig:scatter-comparison}
\end{figure}

Figure~\ref{fig:scatter-comparison} presents a scatter plot of \textit{Ash} versus \textit{Vegetation} percentages within the bounding boxes enclosing all fire clusters in a sample source image and the corresponding target image. Each green dot represents an \textit{Ash-Vegetation} percentage pair from the source image, while each yellow dot represents a corresponding pair from the target image. This comparison illustrates the distributions of \textit{Ash-Vegetation} compositions in the source and target images, with overlapping points indicating similar local compositions.

The upper-right inset of Figure~\ref{fig:scatter-comparison} illustrates the optimal paste-location selection for a single source fire cluster, obtained from Eq.~\eqref{eq:optimal-paste-loc} by minimizing the Euclidean-distance cost function defined in Eq.~\eqref{eq:ash-veg-cost}. The blue point in the inset represents the \textit{Ash-Vegetation} composition of the source fire cluster bounding box $\mathcal{B}_{i,j}$, the red points represent the compositions of the valid candidate regions in the target mask $M_k$ corresponding to $(x,y)\in\mathcal{C}_{i,j,k}$, and the green star represents the composition of the candidate region at the optimal paste location $(x^*,y^*)$. After each fire cluster placement, the cost function $J(x,y)$ is recomposed using the updated target image-mask pair to determine the optimal paste location for the next fire cluster.

% visual image of caccpa results

\subsection{Semantic Segmentation Procedure} 
\label{Method: Semantic Segmentation Procedure}

We implement a U-Net architecture to generate segmentation predictions for wildland fire images and to compare the effectiveness of the aforementioned data augmentation strategies: Rotation-based Augmentation (Section~\ref{Data Augmentation: Rotation}), Non-Context-Aware CCPDA (Section~\ref{Data Augmentation: Non-Context-Aware}), and Context-Aware CCPDA (Section~\ref{Data Augmentation: Context-Aware}). Originally developed for biomedical image segmentation~\cite{ronneberger2015u}, U-Net has since become a widely adopted model for pixel-wise classification due to its U-shaped encoder-decoder architecture with skip connections, which enables efficient feature extraction and precise semantic localization. The U-Net architecture used in this study is illustrated in Figure~\ref{fig:unet_structure}.

\subsection{Evaluation Process for Augmentation Methods} \label{Method: Evaluation of Methods}

A weighted sum-based multi-objective optimization (MOO) approach is employed to evaluate segmentation performance across the different augmentation methods. This approach combines multiple competing metrics into a single score using a linear weighting scheme, assigning high weights to more important metrics while still accounting for less important ones~\cite{chakraborty2023moo, marler2010weighted}. Our focus is on facilitating wildfire management by minimizing missed wildfire detections and improving fire spread prediction. This objective motivates the selection of \textit{Fire}-FNR, \textit{Vegetation}-IoU, and total-IoU as key evaluation metrics~\cite{kim2026ccpda}. The IoU and FNR are defined as:
\begin{equation}
    \notag 
    \text{IoU} = \frac{TP}{TP + FP + FN}, \qquad
    \text{FNR} = \frac{FN}{TP + FN},
\end{equation}
where $TP$, $FP$, and $FN$ denote true positives, false positives, and false negatives, respectively. The weighted-sum score is computed as a linear combination of these three metrics, where each metric is assigned a weight $w_i$ based on the rank order centroid (ROC) method~\cite{ahn2011roc}. The scoring function is defined as:
\begin{align}
    F(x) &= w_1 (1 - f_{\text{\textit{Fire}-FNR}}(x)) \nonumber \\
         &\quad + w_2 f_{\text{\textit{Vegetation}-IoU}}(x) + w_3 f_{\text{total-IoU}}(x).  \label{eq:score}
\end{align}
The weight for each metric is computed as follows:
\begin{equation}
    w_i = \frac{1}{n} \sum_{k=i}^{n} \frac{1}{k},
    \label{eq:roc}
\end{equation}
where $n$ is the number of metrics and $k$ is the rank of each metric. Given that we use three metrics, the resulting weights are $w_1 = 0.611$, $w_2 = 0.278$, and $w_3 = 0.111$. Minimizing missed wildfire detections is critical for field operations, vegetation mapping supports fire spread prediction, and total-IoU provides general situational awareness and map reliability for wildland fire management personnel.

\subsection{Hyperparameter Tuning} 
\label{Method: Hyperparameter Tuning} 

To optimize model performance prior to segmentation experiments, we conducted a comprehensive hyperparameter tuning study. The objective was to identify a model configuration that minimizes \textit{Fire}-FNR while maintaining competitive performance on \textit{Vegetation}-IoU and total-IoU metrics. To ensure fairness and consistency, tuning was performed using the same composite scoring function $F(x)$ introduced in Section~\ref{Method: Evaluation of Methods}, with the \textit{Fire} class prioritized through the rank order centroid (ROC) weighting scheme. We evaluated the effects of learning rate, dropout rate, and batch size on segmentation performance using a grid search across 48 configurations. Each configuration was trained for 80 epochs with the Adam optimizer and categorical cross-entropy loss.

\subsubsection{Search Space and Evaluation Metrics}
\label{HT: Search Space and Evaluation Metrics}

The following values were used:
\begin{itemize}
\item Learning rate: {0.01, 0.005, 0.001, 0.0005}
\item Dropout rate: {0.0, 0.1, 0.2, 0.3}
\item Batch size: {4, 8, 16}
\end{itemize}
The full grid search resulted in $4 \times 4 \times 3 = 48$ unique model configurations. The $F(x)$ score, described in Eq.~\eqref{eq:score}, was used to rank all configurations, with \textit{Fire}-FNR, \textit{Vegetation}-IoU, and total-IoU reported for interpretability. Table~\ref{tab:tuning-summary} lists the three highest and three lowest scoring configurations, with the first row indicating the final hyperparameters used throughout this study.

\subsubsection{Optimal Configuration and Analysis}
\label{HT: Optimal Configuration and Analysis}

\begin{table}[!htb]
\centering
\renewcommand{\arraystretch}{1.2}
\begin{tabular}{@{}cccc@{}}
\toprule
Learning Rate & Dropout Rate & Batch Size & $F(x)$ \\
\midrule
\textbf{0.0005} & \textbf{0.3} & \textbf{4}  & \textbf{0.83734} \\
0.0005  & 0.2 & 16  & 0.83650 \\
0.0005  & 0.2 & 4  & 0.83570 \\
\multicolumn{4}{c}{\hspace{1.2cm}$\vdots$} \\
0.005 & 0.3 & 8  & 0.23578 \\
0.01 & 0.1 & 4  & 0.23578 \\
0.01 & 0.1 & 16  & 0.23578 \\
\bottomrule
\end{tabular}
\captionsetup{justification=centering}
\caption{Selected hyperparameter configurations ranked by $F(x)$ score. The first row, shown in bold, indicates the best-performing tuned configuration.}
\label{tab:tuning-summary}
\end{table}

As shown in Table~\ref{tab:tuning-summary}, although the $F(x)$ scores among the top configurations differ only marginally, their implications for fire detection accuracy are substantial. The best-performing configuration (learning rate = 0.0005, dropout = 0.3, batch size = 4) achieved a $F(x)$ score of 0.83734. In contrast, the lowest-ranked configurations, despite obtaining similar IoU values, suffered from a complete failure to detect \textit{Fire} pixels, resulting in $F(x)$ scores as low as 0.23578. This discrepancy highlights the model sensitivity to hyperparameter choice, particularly, dropout rate and learning rate, when segmenting \textit{Fire} pixels. 

The Study 1 \textit{Fire} [\%] column of Table~\ref{table:results_pixel_pct} indicates that \textit{Fire} pixels constitute only about 5-11\% of the total pixels in the training data across the methods evaluated in Section~\ref{sec:Study01}, making the model particularly sensitive to hyperparameter settings when learning this minority class. Fire clusters are often small, irregularly shaped, and easily confused with \textit{Background} or \textit{Vegetation} due to similar visual characteristics. Consequently, even minor changes in regularization strength or optimization dynamics can severely impair the model’s ability to learn discriminative fire features.

These findings reaffirm the importance of using a fire-sensitive scoring function such as $F(x)$, which explicitly prioritizes low \textit{Fire}-FNR while still accounting for performance in other classes. This approach ensures that hyperparameter tuning remains aligned with the overarching goal of minimizing missed wildfire detections. Consequently, the top-ranked configuration (learning rate = 0.0005, dropout = 0.3, batch size = 4) was adopted for all subsequent experiments in this study. Its ability to balance fire segmentation accuracy with overall robustness provides a reliable foundation for evaluating the effects of different data augmentation strategies.

% RESULTS
\section{Experimental Results}
\label{Results}
% Intro Paragraph 
In this section, we conduct a series of comparative model performance studies using the weighted sum-based scoring method described in Section~\ref{Method: Evaluation of Methods}. The first numerical study concerns the primary objective of this work: determining whether the Context-Aware CCPDA method outperforms other data augmentation strategies for multiclass wildland fire image segmentation using the U-Net model. Following that, we examine whether the same erosion level provides the best performance for the proposed Context-Aware CCPDA method before and after expansion of the original image set. Specifically, we apply the Context-Aware CCPDA method at erosion levels of 0\%, 10\%, 20\%, and 30\% to two image sets: the original 20-image set of non-augmented RGB images and corresponding ground-truth labels used to construct the BURN 1~\cite{BURN01} dataset, and the expanded 26-image set used to construct the BURN 2~\cite{BURN02} dataset. This study determines whether the best-performing erosion level remains consistent even after expansion of the original image set and whether retaining the centralized core regions of fire clusters remains effective under the Context-Aware CCPDA framework. Finally, we extend the comparison of segmentation performance beyond the U-Net model and compare the impact of augmentation methods on additional segmentation architectures, including FCN~\cite{long2015fcn}, SegNet~\cite{badrinarayanan2017segnet}, and DeepLabV3+~\cite{chen2018deeplabv3+}. This final experiment evaluates whether the observed performance trends are consistent across multiple multiclass segmentation models or if the performance gain is limited to the U-Net architecture. Thus, the studies are organized as follows:
\begin{itemize}
\item \textbf{Study 1}: Comparison of multiclass fire scene segmentation by training the U-Net model using data from four augmentation methods: Non-Augmented (baseline, Method 1), Rotation-by-18$^\circ$ (Method 2), Non-Context-Aware Copy-Paste (Method 3), and Context-Aware Copy-Paste (Method 4).
\item \textbf{Study 2}: Evaluation of the Context-Aware CCPDA method across erosion levels of 0\%, 10\%, 20\%, and 30\% using the original 20-image set and the expanded 26-image set.
\item \textbf{Study 3}: Comparison of the four augmentation methods from Study 1 across three benchmark deep-learning semantic segmentation models: FCN, SegNet, and DeepLabV3+.
\end{itemize}

% STUDY 01
\subsection{Study 1: U-Net Comparison of Augmentation Methods} \label{sec:Study01}

Study 1 compares the impact on U-Net based segmentation performance of four methods: Non-Augmented (baseline, Method 1), Rotation-by-18$^\circ$ (Method 2), Non-Context-Aware Copy-Paste (Method 3) and Context-Aware Copy-Paste (Method 4). The goal is to evaluate whether contextually correct data augmentation (Method 4) enables more accurate segmentation compared to contextually agnostic methods (Methods 1-3). For a controlled comparison, all methods in Study 1 are evaluated at an erosion level of 0\%, so that differences in segmentation performance can be attributed to the augmentation methods rather than to the erosion level.

\begin{figure*}[!ht]
    \centering
    \captionsetup{justification=centering}
    \includegraphics[width=0.95\textwidth]{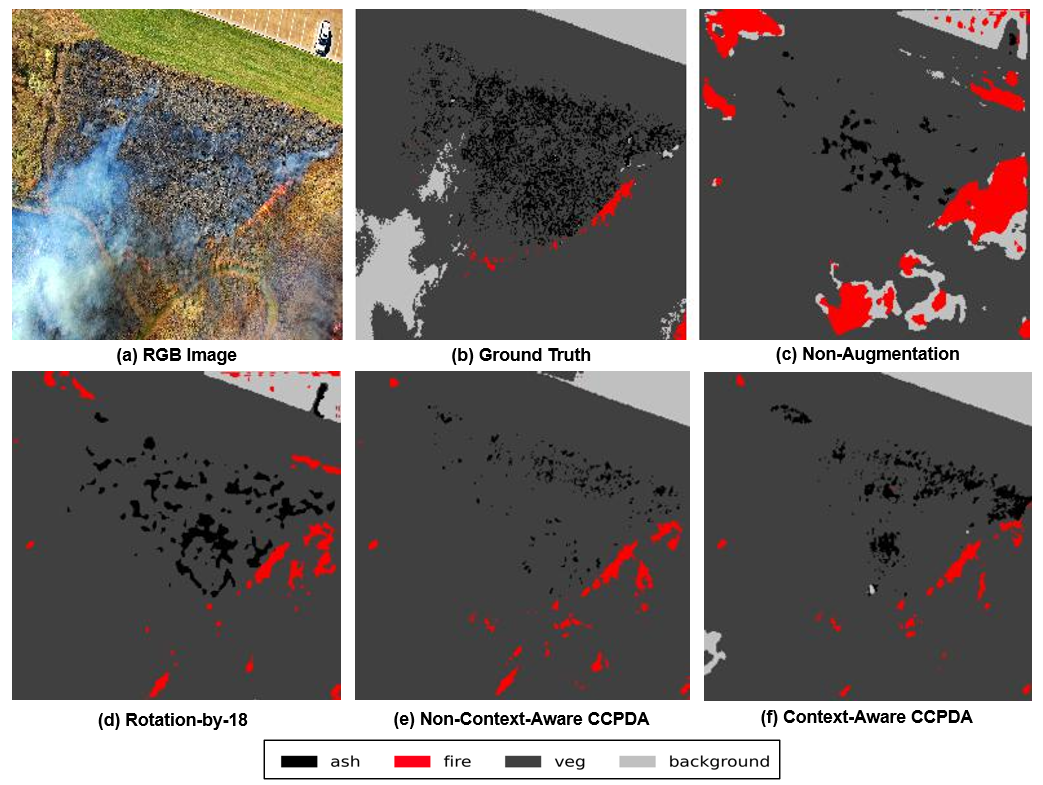}
    \caption{A visual comparison of the semantic segmentation results of Study 1 for a sample test RGB image.}
    \label{fig:visual_study01}
\end{figure*}

A visual sample of segmentation predictions for the four augmentation methods is shown in Figure~\ref{fig:visual_study01}, along with the original RGB test image and the corresponding ground truth annotations. Focusing on the \textit{Fire} class, we observe that all methods except the Context-Aware CCPDA approach exhibit noticeable over-prediction and misclassification. The non-CCPDA approaches (Methods 1–2) misclassify some non-burnable regions, including portions of the parking lot, as \textit{Fire} rather than \textit{Background}. These methods may cause the U-Net to associate the parking lot surface with the \textit{Fire} class. This effect is likely reinforced by the relatively high red-channel intensity in the parking lot region of the RGB image in Figure~\ref{fig:visual_study01}, which makes the region visually similar to fire pixels and may contribute to misclassification. In contrast, U-Net models trained with CCPDA-based methods (Methods 3–4) produce more precise segmentation results, more effectively capturing the complex boundaries of fire regions. Although minor over-prediction persists in the context-aware augmentation strategy (Method 4), it remains less pronounced than in the non-context-aware approach (Method 3). Such an over-prediction is generally less concerning than missed detections of fire in wildland fire management.

In addition, segmentation predictions for the other three classes---\textit{Ash}, \textit{Vegetation}, and \textit{Background} generally improve going from the non-augmented and rotation-based methods to the CCPDA-based methods. Examining the heavy smoke region in the lower left corner (classified as \textit{Background}), we observe that only the Context-Aware CCPDA method (Method 4) correctly predicts part of this region as \textit{Background}. Additionally, the white vehicle in the parking lot is misclassified as \textit{Ash} or \textit{Vegetation} by the non-CCPDA methods (Methods 1–2), while CCPDA-based methods (Methods 3–4) correctly label it as \textit{Background}, despite the vehicle’s distinctive color and shape relative to the surrounding background. None of the methods accurately predict the sparse ash regions concentrated in the center. Although the Context-Aware CCPDA strategy (Method 4) performs slightly better than the other methods, \textit{Ash} pixels are frequently misclassified as \textit{Vegetation} across all methods. This limitation likely arises because ash regions are often intermixed with vegetation and exhibit similar visual characteristics in RGB imagery, making them difficult for the model to distinguish reliably. Overall, based on visual analysis, the Context-Aware CCPDA method produces the most accurate segmentation relative to ground truth across all four methods.

Numerical quantification of segmentation accuracy is presented in Table~\ref{table:study01_comparison}, which aligns with the visual results shown. In particular, \textit{Fire}-FNR progressively decreases from Methods~(1)-(4). The Context-Aware CCPDA approach achieves the lowest \textit{Fire}-FNR at 6.75\%, compared to 8.24\% for the Non-Context-Aware CCPDA method, corresponding to a relative improvement of 18.08\%. This suggests that context-guided placement of fire clusters effectively supports our primary objective of minimizing missed wildfire detections.

Finally, the weighted-sum score $F(x)$ shows that the Context-Aware CCPDA method (Method 4) achieves the highest value of 0.8458, representing a 2.26\% improvement over the Non-Context-Aware CCPDA method (Method 3), which achieves a score of 0.8271. This improvement is accompanied by moderate gains in \textit{Vegetation}-IoU and total-IoU between the two methods. In particular, the Study 1 \textit{Fire} [\%] column of Table~\ref{table:results_pixel_pct} shows that the proportion of \textit{Fire} pixels in the augmented training data decreases from the Rotation-by-18$^\circ$ method to the copy-paste methods (Methods 3-4). Despite this lower proportion of \textit{Fire} pixels in the training data, the Context-Aware CCPDA method achieves the best segmentation performance, indicating that contextually accurate augmentation can improve class learning beyond simply increasing pixel counts. Indeed, by placing copied fire clusters in contextually correct locations, the model achieves a meaningful reduction in \textit{Fire}-FNR while maintaining strong overall segmentation performance.

% study01 table
\begin{table}[t]
\centering
\renewcommand{\arraystretch}{1.2}
\resizebox{\columnwidth}{!}{%
\begin{tabular}{>{\centering\arraybackslash}m{3cm} c c c c}
    \hline
    Method & \textit{Fire}-FNR & \textit{Veg.}-IoU & total-IoU & $F(x)$ \\
    \hline
    Non-Augmented & 18.04 & 67.40 & 54.45 & 0.7486\\
    Rotation-by-18$^\circ$ & 14.17 & 71.14 & 60.54 & 0.7894\\
    Non-Context-Aware Copy-Paste & 8.24 & 71.52 & 60.91 & 0.8271\\
    \textbf{Context-Aware Copy-Paste} & \textbf{6.75} & \textbf{73.74} & \textbf{64.03} & \textbf{0.8458} \\
    \hline
\end{tabular}%
}
\captionsetup{justification=centering}
\caption{Study 1 performance metrics (\%) and weighted-sum scores across different augmentation methods.}
\label{table:study01_comparison}
\end{table}

% study01 and study02 stats table 
\begin{table}[t]
\centering
\renewcommand{\arraystretch}{1.2}
\resizebox{\columnwidth}{!}{%
\begin{tabular}{>{\centering\arraybackslash}m{3.1cm} c | c c}
    \hline
    (Study 1) Method & \textit{Fire} [\%] & (Study 2) Erosion \% & \textit{Fire} [\%]\\
    \hline
    Non-Augmented & 5.07 & 0\% & 8.80\\
    Rotation-by-18$^\circ$ & 10.59 & 10\% & 8.31\\
    N.C.A$^{1}$ Copy-Paste & 9.30 & 20\% & 7.79\\
    C.A$^{2}$ Copy-Paste & 9.19 & 30\% & 7.38\\
    \hline
    \multicolumn{3}{c}{Non-Augmented Data Pixel Total} & 655,360\\
    \multicolumn{3}{c}{Augmented Data Pixel Total} & 13,107,200\\
    \hline
\end{tabular}%
}
\captionsetup{justification=centering}
\caption{Comparison of \textit{Fire} pixel percentages across Study 1 augmentation methods and Study 2 erosion levels for the expanded 26-image set, including total pixel counts for the corresponding non-augmented and augmented training datasets. $^{1}$Non-Context-Aware, $^{2}$Context-Aware.}
\label{table:results_pixel_pct}
\end{table}

% visual image of study01

% STUDY 02
\subsection{Study 2: Performance Evaluation of the Context-Aware CCPDA Method Across Erosion Levels} 
\label{sec:Study02}

Study 2 examines the segmentation performance of Context-Aware CCPDA  across four erosion levels (0\%, 10\%, 20\%, and 30\%) using the original 20-image set and the expanded 26-image set. The objective is to determine whether the same erosion level provides the best performance for the Context-Aware CCPDA method before and after expansion of the original image set. This comparison also examines how changes in erosion level affect \textit{Fire}-FNR and overall segmentation performance across the two image sets.

\begin{figure*}[!ht]
    \centering
    \captionsetup{justification=centering}
    \includegraphics[width=0.95\textwidth]{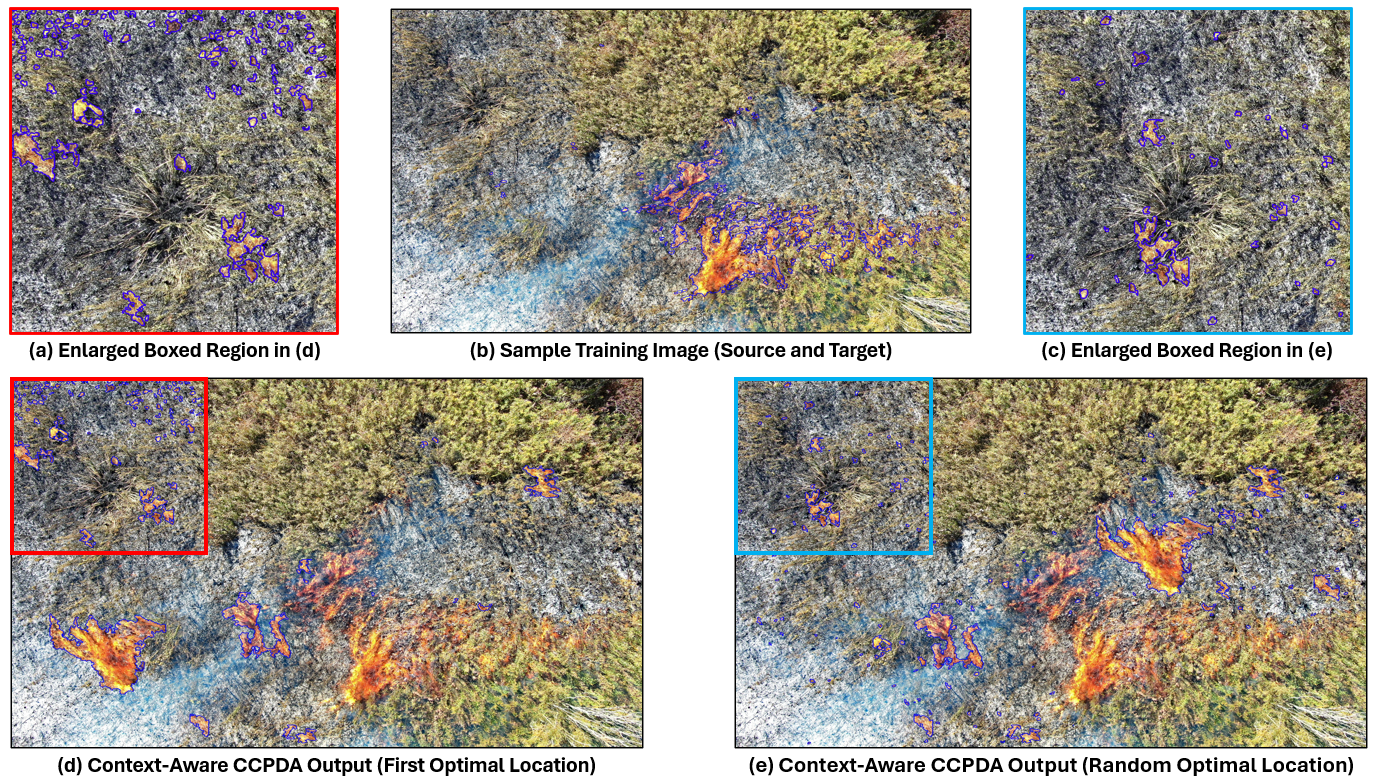}
    \caption{A visual comparison of Context-Aware CCPDA outputs using the first and a randomly selected optimal location for a sample training image. Sub-figure (b) shows the training image used as both the source and target image. Sub-figures (d) and (e) show the Context-Aware CCPDA outputs obtained by pasting the source fire clusters at the first optimal location and at a randomly selected location from the set of possible optimal locations in the target image, respectively. Sub-figures (a) and (c) provide enlarged views of the corresponding boxed regions in (d) and (e), respectively.}
    \label{fig:nonrand_vs_random_ca_ccpda}
\end{figure*}

For this study, we use a controlled variant of the Context-Aware CCPDA method. The original Context-Aware CCPDA method, described in Section~\ref{Data Augmentation: Context-Aware}, randomly selects among multiple optimal paste locations. In the controlled variant, when multiple optimal paste locations are identified, the candidate locations are evaluated from left to right and top to bottom, and the first optimal location encountered is selected instead of randomly selecting among the optimal locations. As illustrated in Figure~\ref{fig:nonrand_vs_random_ca_ccpda}(a) and (d), this controlled selection results in fire clusters being more concentrated toward the upper-left region of the target image, whereas the original random-selection approach shown in Figure~\ref{fig:nonrand_vs_random_ca_ccpda}(c) and (e) produces more spatially distributed placements. This procedure ensures consistent, context-aware paste-location selection across the tested erosion levels, so that erosion level remains the primary varying factor.

Per Table~\ref{table:study02_comparison}, 10\% erosion level yields the highest overall performance for both the original 20-image set and the expanded 26-image set. For the 20-image set, increasing the erosion level from 0\% to 10\% reduces \textit{Fire}-FNR from 5.22\% to 4.00\% and increases the weighted-sum score from 0.8276 to 0.8365. Similarly, for the 26-image set, \textit{Fire}-FNR decreases from 6.76\% to 4.67\%, while the weighted-sum score increases from 0.8323 to 0.8460. However, further increasing the erosion level to 20\% and 30\% degrades segmentation performance of both image sets, as reflected by progressively higher \textit{Fire}-FNR values and lower $F(x)$ scores relative to the 10\% erosion level. These results show that the best-performing erosion level remains consistent across both image sets under the Context-Aware CCPDA framework. In addition, this is consistent with the best-performing setting identified in our previous Non-Context-Aware CCPDA study~\cite{kim2026ccpda}.

The above results underscore the challenge of manual labeling at fire cluster boundaries, where fire and vegetation often appear visually similar. Even with accurate annotations, the model may misclassify pixels due to similar visual features. Eroding each fire cluster by 10\% helps eliminate ambiguous boundary pixels that could introduce noise during training. However, increasing the erosion level beyond this point causes segmentation performance to decline, likely due to the loss of accurately labeled \textit{Fire} pixels. % essential for model training. 
This trend is further supported by the Study 2 \textit{Fire} [\%] column of Table~\ref{table:results_pixel_pct}, which shows that the proportion of \textit{Fire} pixels in the augmented training data decreases steadily from 8.80\% at 0\% erosion to 7.38\% at 30\% erosion. Considered together with the results in Table~\ref{table:study02_comparison}, this suggests that excessive erosion removes an increasing number of \textit{Fire} pixels that could otherwise contribute useful information during training. Therefore, we conclude that the Context-Aware Copy-Paste approach with a 10\% erosion level provides the best segmentation performance among the tested erosion levels for both the original 20-image set and the expanded 26-image set.

% Study 2 table (updated)
\begin{table*}[t]
\centering
\renewcommand{\arraystretch}{1.2}
\resizebox{\textwidth}{!}{%
\begin{tabular}{>{\centering\arraybackslash}m{3cm} c c c c | c c c c}
    \hline
    & \multicolumn{4}{c|}{\textbf{20-image set}} 
    & \multicolumn{4}{c}{\textbf{26-image set}} \\
    \cline{2-9}
    Method 
    & \textit{Fire}-FNR & \textit{Veg.}-IoU & total-IoU & $F(x)$
    & \textit{Fire}-FNR & \textit{Veg.}-IoU & total-IoU & $F(x)$ \\
    \hline

    Erosion-by-0\%
    & 5.22 & 66.98 & 56.16 & 0.8276
    & 6.76 & 70.32 & 60.45 & 0.8323 \\

    \textbf{Erosion-by-10\%}
    & \textbf{4.00} & \textbf{66.79} & \textbf{57.93} & \textbf{0.8365}
    & \textbf{4.67} & \textbf{70.78} & \textbf{60.17} & \textbf{0.8460} \\

    Erosion-by-20\%
    & 7.59 & 67.99 & 57.29 & 0.8172
    & 8.28 & 70.34 & 58.88 & 0.8213 \\

    Erosion-by-30\%
    & 9.04 & 68.15 & 57.09 & 0.8086
    & 12.27 & 72.31 & 62.05 & 0.8059 \\

    \hline
\end{tabular}%
}
\captionsetup{justification=centering}
\caption{Study 2 performance metrics (\%) and weighted-sum scores across erosion levels for the Context-Aware CCPDA method on the original 20-image set used to construct BURN 1 and the expanded 26-image set used to construct BURN 2.}
\label{table:study02_comparison}
\end{table*}

\subsection{Study 3: Comparison Across Deep Learning Segmentation Architectures} 
\label{sec:Study03}

Study 3 evaluates the performance of the four data augmentation strategies---Non-Augmented (baseline, Method 1), Rotation-by-18$^\circ$ (Method 2), Non-Context-Aware Copy-Paste (Method 3), and Context-Aware Copy-Paste (Method 4)---across three benchmark semantic segmentation models: FCN, SegNet, and DeepLabV3+. The objective is to determine whether the CCPDA methods generalize beyond the U-Net model; in other words, whether the performance trends observed in Study 1 remain consistent across different architectures. Examining Table~\ref{table:study03_comparison}, we observe that all three models exhibit trends consistent with those observed for the U-Net model in Study 1. In particular, \textit{Fire}-FNR progressively decreases from Methods~(1)-(4) across all models, while the Context-Aware CCPDA method consistently achieves the highest $F(x)$ score among the augmentation methods.

For example, SegNet shows a noticeable improvement when the Non-Context-Aware and Context-Aware CCPDA methods are compared, with \textit{Fire}-FNR decreasing from 10.17\% to 7.78\% (a 23.50\% reduction) and the $F(x)$ score increasing from 0.8180 to 0.8311 (a 1.60\% improvement). In contrast, FCN exhibits the same trend but only modest gains, with \textit{Fire}-FNR decreasing from 22.54\% to 21.72\% (a 3.64\% reduction) and the $F(x)$ score increasing from 0.7208 to 0.7211 (a 0.04\% improvement). The relatively small improvement observed for FCN is likely due to its simpler architecture, which relies on coarse upsampling and skip fusion rather than a full decoder to reconstruct fine spatial details of the dataset. In contrast, models such as SegNet and DeepLabV3+ employ more advanced encoder-decoder structures that better preserve spatial information, enabling them to more effectively leverage context-aware augmentation for improved wildland fire image segmentation. Overall, these results indicate that the Context-Aware CCPDA method consistently achieves the lowest \textit{Fire}-FNR and the highest $F(x)$ score across different deep learning segmentation architectures, reinforcing its effectiveness as a general data augmentation strategy for the segmentation of wildland fire scenes.

% study03 stats table
\begin{table}[t]
\centering
\renewcommand{\arraystretch}{1.2}
\resizebox{\columnwidth}{!}{%
\begin{tabular}{>{\centering\arraybackslash}m{3cm} c c c c}
    \hline
    Method & \textit{Fire}-FNR & \textit{Veg.}-IoU & total-IoU & $F(x)$ \\
    \hline
    \multicolumn{5}{c}{\textbf{FCN}} \\
    Non-Augmented & 72.85 & 65.23 & 49.72 & 0.4024 \\
    Rotation-by-18$^\circ$ & 25.35 & 66.82 & 54.67 & 0.7026 \\
    Non-Context-Aware Copy-Paste & 22.54 & 67.33 & 54.34 & 0.7208 \\
    \textbf{Context-Aware Copy-Paste} & \textbf{21.72} & \textbf{66.19} & \textbf{52.99} & \textbf{0.7211} \\
    \hline
    \multicolumn{5}{c}{\textbf{SegNet}} \\
    Non-Augmented & 40.45 & 1.22 & 2.18 & 0.3697 \\
    Rotation-by-18$^\circ$ & 13.34 & 71.95 & 60.99 & 0.7972 \\
    Non-Context-Aware Copy-Paste & 10.17 & 72.32 & 61.36 & 0.8180 \\
    \textbf{Context-Aware Copy-Paste} & \textbf{7.78} & \textbf{71.75} & \textbf{61.38} & \textbf{0.8311} \\
    \hline
    \multicolumn{5}{c}{\textbf{DeepLabV3+}} \\
    Non-Augmented & 100.00 & 49.51 & 29.08 & 0.1699 \\
    Rotation-by-18$^\circ$ & 12.51 & 71.68 & 60.24 & 0.8007 \\
    Non-Context-Aware Copy-Paste & 11.04 & 70.08 & 58.92 & 0.8038 \\
   \textbf{Context-Aware Copy-Paste} & \textbf{9.56} & \textbf{70.44} & \textbf{59.64} & \textbf{0.8146} \\
    \hline
\end{tabular}%
}
\captionsetup{justification=centering}
\caption{Study 3 performance metrics (\%) and weighted-sum scores across different augmentation methods on three semantic segmentation models: FCN, SegNet, and DeepLabV3+.}
\label{table:study03_comparison}
\end{table}

% CONCLUSION
\section{Conclusion}
\label{Conclsuion}

This paper presented the Centralized Copy-Paste Data Augmentation (CCPDA) method by incorporating contextual information, thereby improving the realism and accuracy of wildland fire image segmentation under data-scarce conditions. The Context-Aware CCPDA %approach presented in this paper 
guides the placement of fire clusters based on the \textit{Ash–Vegetation} composition of local regions, with the goal of minimizing false negatives and preserving semantic fidelity during training.

The Context-Aware CCPDA method improves wildland fire semantic segmentation by placing copied fire clusters in contextually appropriate regions while avoiding overlap with existing \textit{Fire} pixels and non-burnable \textit{Background} regions. In Study 1, comparison across methods showed that the Context-Aware CCPDA method achieved the lowest \textit{Fire}-FNR of 6.75\% and the highest weighted-sum score of $F(x)=0.8458$, demonstrating improved the \textit{Fire} class detection while maintaining strong overall segmentation performance. Next, in Study 2, the erosion-level analysis showed that 10\% erosion provided the best performance among the tested erosion levels for both the original 20-image set and the expanded 26-image set, indicating that the best-performing erosion level remained consistent even after expansion of the original annotated image set. Finally, in Study 3, the Context-Aware CCPDA method consistently achieved the lowest \textit{Fire}-FNR and highest $F(x)$ score among the tested methods across FCN, SegNet, and DeepLabV3+, demonstrating that the observed performance improvements are not limited to the U-Net architecture. Collectively, these results demonstrate the effectiveness and generalizability of the proposed Context-Aware CCPDA method for multiclass wildland fire image segmentation under limited labeled data conditions.

One direction for future work is to extend the Context-Aware CCPDA method to dynamic wildfire monitoring in video data, which will require a temporally consistent segmentation procedure. Applying this method to sequential frames could enable real-time inference and early warning systems for in-field fire response~\cite{duangsuwan2023accuracy, kim2015real}. In addition, the robustness of our approach to smoke-induced occlusion remains an open question, as smoke often obscures fire regions and degrades segmentation performance in wildfire imagery~\cite{ma2025semi, chaturvedi2022survey}. Addressing this challenge will further improve the effectiveness of the Context-Aware CCPDA data augmentation strategy and enhance semantic segmentation accuracy for real-world wildfire monitoring.

% Acknowledgment
\section{Acknowledgment}
The authors thank Nicholas Sansoterra, a graduate student at The Ohio State University, for his assistance in implementing the benchmark testing code for evaluating the augmentation methods across multiple semantic segmentation architectures.
The authors also thank Dana Costa, Kewei Du, Alex Guller, and Stanley Zachariah, undergraduate students at The Ohio State University, for their assistance in annotating the expanded BURN 2 dataset comprising 26 images.

% Bibliography 
{\small
\bibliographystyle{IEEEtran}
\bibliography{IEEEexample}
}
\end{document}